\documentclass[runningheads]{llncs}
\usepackage{makeidx}
\usepackage{graphicx}
\usepackage{amsmath,amssymb} %
\usepackage{color}
\usepackage{listings}
\usepackage{booktabs} %
\usepackage{subcaption}
\usepackage{subfig}
\usepackage{rotating} %
\usepackage{multirow}

\usepackage{xspace}
 \makeatletter
 \DeclareRobustCommand\onedot{\futurelet\@let@token\@onedot}
 \def\@onedot{\ifx\@let@token.\else.\null\fi\xspace}
 \def\eg{e.g\onedot} \def\Eg{E.g\onedot}
 \def\ie{i.e\onedot}

 \makeatother

\DeclareRobustCommand{\Figref}[1]{Figure~\ref{#1}}
\DeclareRobustCommand{\Figsref}[1]{Figures~\ref{#1}}

\DeclareRobustCommand{\Secref}[1]{Section~\ref{#1}}
\DeclareRobustCommand{\secref}[1]{Section~\ref{#1}}

\DeclareRobustCommand{\Tableref}[1]{Table~\ref{#1}}

\DeclareRobustCommand{\Tablesref}[1]{Tables~\ref{#1}}

\newcommand{\invisible}[1]{}%

\graphicspath{{./fig/}{./fig/plots/}}

\begin{document}
\pagestyle{headings}
\mainmatter

\title{The Long-Short Story of Movie Description}
\titlerunning{The Long-Short Story of Movie Description.}
\authorrunning{Anna Rohrbach \and Marcus Rohrbach \and Bernt Schiele.}
\author{Anna Rohrbach$^{1}$ \and Marcus Rohrbach$^{2}$ \and Bernt Schiele$^{1}$}
\institute{Max Planck Institute for Informatics, Saarbr{\"u}cken, Germany \and
           UC Berkeley EECS and ICSI, Berkeley, CA, United States}
\maketitle

\begin{abstract}
Generating descriptions for videos has many applications including assisting blind people and human-robot interaction. The recent advances in image captioning as well as the release of large-scale movie description datasets such as MPII-MD \cite{rohrbach15cvpr} allow to study this task in more depth. 
Many of the proposed methods for image captioning rely on pre-trained object classifier CNNs and Long-Short Term Memory recurrent networks (LSTMs) for generating descriptions. 
While image description focuses on objects, we argue that it is important to distinguish verbs, objects, and places in the challenging setting of movie description. In this work we show how to learn robust visual classifiers from the weak annotations of the sentence descriptions. Based on these visual classifiers we learn how to generate a description using an LSTM.
We explore different design choices to build and train the LSTM and achieve the best performance to date on the challenging MPII-MD dataset. We compare and analyze our approach and prior work along various dimensions to better understand the key challenges of the movie description task.
\end{abstract}

\section{Introduction}
\label{sec:intro}

Automatic description of visual content has lately received a lot of interest in our community. Multiple works have successfully addressed the image captioning problem \cite{donahue15cvpr,karpathy15cvpr,kiros15tacl,vinyals15cvpr}. %
Many of the proposed methods rely on %
Long-Short Term Memory networks (LSTMs) \cite{hochreiter97nc}. %
In the meanwhile, two large-scale movie description datasets have been proposed, namely MPII Movie Description (MPII-MD) \cite{rohrbach15cvpr} and  Montreal Video Annotation Dataset (M-VAD) \cite{torabi15arxiv}. Both are based on movies with associated textual descriptions and allow studying the problem how to generate movie description for visually disabled people. %
Works addressing these datasets \cite{rohrbach15cvpr,venugopalan15arxiv,yao15arxiv} show that they are indeed challenging in terms of visual recognition and automatic description. This results in %
 a significantly lower performance %
 then on simpler video datasets (\eg MSVD \cite{chen11acl}), %
but a detailed analysis of the difficulties is missing. In this work we address this by taking a closer look at the performance of existing methods on the movie description task. 

This work contributes a) an approach to build robust visual classifiers  which distinguish verbs, objects, and places extracted from weak sentence annotations;     
 b) based on the visual classifiers we evaluate different design choices to train an LSTM for generating descriptions. This outperforms related work on the MPII-MD dataset, both using automatic and human evaluation; c) we perform a detailed analysis of prior work and our approach to understand the challenges of the movie description task.

\section{Related Work}
\label{sec:related}

\paragraph{Image captioning.}
Automatic image description has been studied in the past \cite{farhadi10eccv,kulkarni11cvpr,kuznetsova14tacl,mitchell12eacl}, however it regained attention just recently. Multiple works have been proposed \cite{donahue15cvpr,fang15cvpr,karpathy15cvpr,kiros15tacl,mao14arXiv,vinyals15cvpr,xu2015show}. Many of them rely on Recurrent Neural Networks (RNNs) and in particular on Long-Short Term Memory networks (LSTMs). Also new datasets have been released, Flickr30k \cite{young2014image} and MS COCO Captions \cite{capeval2015}, where \cite{capeval2015} additionally presents a standardized setup for image captioning evaluation. There are also attempts to analyze the performance of recent methods. \Eg \cite{devlin2015language} compares them with respect to the novelty of generated descriptions and additionally proposes a nearest neighbor baseline that improves over recent methods. %

\paragraph{Video description.}
In the past video description has been addressed in semi-realistic settings \cite{barbu12uai,kojima02ijcv}, on a small scale \cite{das13cvpr,guadarrama13iccv,thomason14coling} or in constrained scenarios like cooking \cite{rohrbach14gcpr,rohrbach13iccv}. 
Most works (with a few exceptions, \eg \cite{rohrbach14gcpr}) study the task of describing a short clip with a single sentence. \cite{donahue15cvpr} first proposed to describe videos using an LSTM, relying on precomputed CRF scores from \cite{rohrbach14gcpr}. \cite{venugopalan15naacl} extended this work to extract CNN features from frames which are max-pooled over time. They show the benefit of pre-training the LSTM network for image captioning and fine-tuning it to video description. 
 \cite{pan2015arxiv} proposes a framework that consists of a 2-D and/or 3-D CNN and the LSTM is trained jointly with a visual-semantic embedding to ensure better coherence between video and text.
\cite{xu2015aaai} jointly addresses the language generation and video/language retrieval tasks by learning a joint embedding model for a deep video model and compositional semantic language model.

\paragraph{Movie description.} 
 Recently two large-scale movie description datasets have been proposed, MPII Movie Description (MPII-MD) \cite{rohrbach15cvpr} and  Montreal Video Annotation Dataset (M-VAD) \cite{torabi15arxiv}. Given that they are based on movies, they cover a much broader domain then previous video description datasets. Consequently they are much more varied and challenging with respect to the visual content and the associated description. They also do not have any additional annotations, as \eg TACoS Multi-Level \cite{rohrbach14gcpr}, thus one has to rely on the weak annotations of the sentence descriptions.
To handle this challenging scenario \cite{yao15arxiv} proposes an attention based model which selects the most relevant temporal segments in a video and incorporates 3-D CNN and generates a sentence using an LSTM. \cite{venugopalan15arxiv} proposes an encoder-decoder framework, where a single LSTM encodes the input video frame by frame and decodes it into a sentence, outperforming \cite{yao15arxiv}.  Our approach for sentence generation is most similar to \cite{donahue15cvpr} and we also rely on their LSTM implementation based on Caffe \cite{jia2014caffe}. However, we analyze different aspects and variants of this architecture for movie description. %
To extract labels from sentences we rely on the semantic parser of \cite{rohrbach15cvpr}, however we treat the labels differently to handle the weak supervision (see \Secref{sec:visual_labels}). We show that this improves over \cite{rohrbach15cvpr} and \cite{venugopalan15arxiv}.

\section{Approach}
\label{sec:approach}
\begin{figure}[t]
\begin{center}
\includegraphics[width=0.9\linewidth]{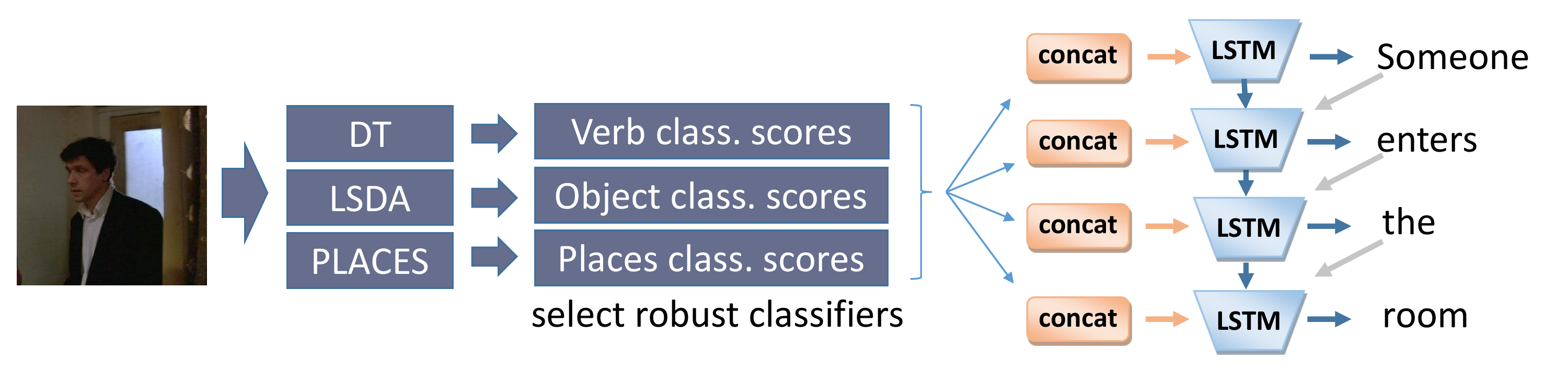}
\caption{Overview of our approach. We first train the visual classifiers for verbs, objects and places, using different visual features: DT (dense trajectories \cite{wang13iccv}), LSDA (large scale object detector \cite{hoffman14nips}) and PLACES (Places-CNN \cite{zhou14nips}). Next, we concatenate the scores from a subset of selected robust classifiers and use them as input to our LSTM.}
\label{fig:approach}
\end{center}
\end{figure}

In this section we present our two-step approach to video description. The first step performs visual recognition, while the second step generates textual descriptions. For the visual recognition we propose to use the visual classifiers trained according to the labels' semantics and ``visuality''. For the language generation we rely on a LSTM network which has been successfully used for image and video description \cite{donahue15cvpr,venugopalan15arxiv}. We discuss various design choices for building and training the LSTM. An overview of our approach is given in \Figref{fig:approach}. 

\subsection{Visual Labels for Robust Visual Classifiers}
\label{sec:visual_labels}
For training we rely on a parallel corpus of videos and weak sentence annotations. As in \cite{rohrbach15cvpr} we parse the sentences to obtain a set of labels (single words or short phrases, \eg \emph{look up}) to train our visual classifiers. However, in contrast to \cite{rohrbach15cvpr} we do not want to keep all of these initial labels as they are noisy, but select only visual ones which actually can be robustly recognized. 
\paragraph{Avoiding parser failure.}
Not all sentences can be parsed successfully, as \eg some sentences are incomplete or grammatically incorrect. To avoid loosing the potential labels in these sentences, we match our set of initial labels to the sentences which the parser failed to process.
\paragraph{Semantic groups.}
Our labels correspond to different semantic groups. In this work we consider three most important groups: verbs (actions), objects and places, as they are typically visual. One could also consider \eg groups like mood or emotions, which are naturally harder for visual recognition. We propose to treat each label group independently. First, we rely on a different representation for the each semantic groups, which is targeted to the specific group. Namely we use the activity recognition feature Improved Dense Trajectories (DT) \cite{wang13iccv} for verbs, large scale object detector responses (LSDA) \cite{hoffman14nips} for objects and scene classification scores (PLACES) \cite{zhou14nips} for places. Second, we train one-vs-all SVM classifiers for each group separately. The intuition behind this is to discard  ``wrong negatives'' (\eg using \emph{object} ``bed'' as negative for \emph{place} ``bedroom''). 
\paragraph{Visual labels.}
Now, how do we  select \emph{visual} labels for our semantic groups? In order to find the verbs among the labels we rely on the semantic parser of \cite{rohrbach15cvpr}. Next, we look up the list of ``places'' used in \cite{zhou14nips} and search for corresponding words among our labels. We look up the object classes used in \cite{hoffman14nips} and search for these ``objects'', as well as their base forms (\eg ``domestic cat'' and ``cat''). We discard all the labels that do not belong to any of our three groups of interest as we assume that they are likely not visual and thus are difficult to recognize. Finally, we discard labels which the classifiers could not learn, as these are likely to be noisy or not visual. 
For this we require the classifiers to have have minimum area under the  ROC-curve (Receiver Operating Characteristic).

\begin{figure}[t]
\begin{center}
\includegraphics[width=12cm]{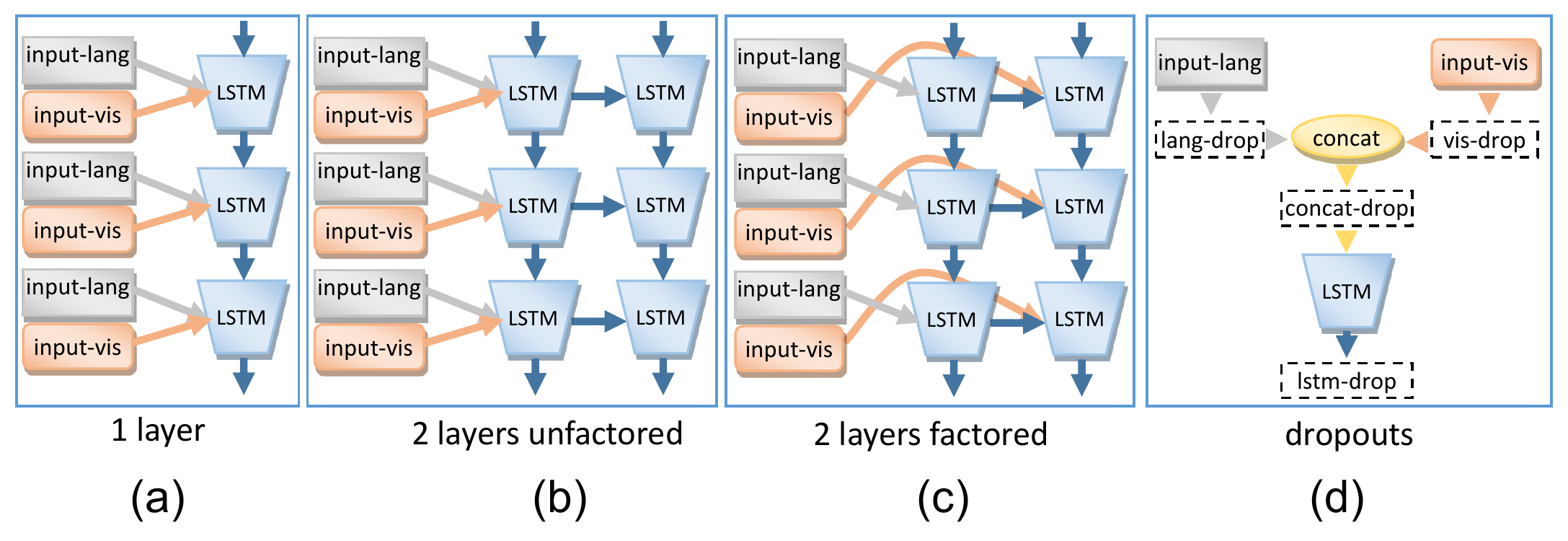}
\end{center}
\vspace{-0.7cm}
\caption{(a-c) LSTM architectures. (d) Variants of placing the dropout layer.}
\vspace{-0.5cm}
\label{fig:lstm_arch}
\end{figure}

\subsection{LSTM for Sentence Generation}
\label{sec:lstm}
We rely on the basic LSTM architecture proposed in \cite{donahue15cvpr} for video description. As shown in \Figsref{fig:approach} and \ref{fig:lstm_arch}(a), at each time step, an LSTM generates a word  and receives the visual classifiers (input-vis)  as well as as the previous generated word (input-lang) as input. To handle natural words we encode each word with a one-hot-vector according to their index in a dictionary and a lower dimensional embedding. The embedding is jointly learned during training of the LSTM. 
\cite{donahue15cvpr}  compares three variants: (a) an encoder-decoder architecture, (b) a decoder architecture with visual max predictions, and (c) a decoder architecture with visual probabilistic predictions.
In this work we rely on variant (c) which was shown to work best as it can rely on the richest visual input. We analyze the following aspects for this architecture:

\paragraph{Layer structure:}
We compare a 1-layer architecture with a 2-layer architecture. In the 2-layer architecture, the output of the first layer is used as input for the second layer (\Figref{fig:lstm_arch}b) and was used by \cite{donahue15cvpr} for video description. Additionally we also compare to a 2-layer factored architecture \cite{donahue15cvpr}, where the first layer only gets the language as input and the second gets the output of the first layer as well as the visual input.
\paragraph{Dropout placement:}
To learn a more robust network which is less likely to overfit we rely on a dropout \cite{hinton2012improving}. Using dropout a ratio $r$ of randomly selected units is set to 0 during training (while all others are multiplied with $1/r$). We explore different ways to place dropout in the network, \ie either for language input (lang-drop) or visual (vis-drop) input only, for both inputs (concat-drop) or for the LSTM output (lstm-drop), see \Figref{fig:lstm_arch}(d). While the default dropout ratio is $r=0.5$, we evaluate the effect of different ratios.
\paragraph{Learning strategy:}
By default we rely on a step-based learning strategy, where a learning rate is halved after a certain number of steps. We find the best learning rate and step size on the validation set. Additionally we compare this to a polynomial learning strategy, where the learning rate is continuously decreased. The polynomial learning strategy has been shown to give good results faster without tweaking step size for GoogleNet implemented by Sergio Guadarrama in Caffe \cite{jia2014caffe}.

\section{Evaluation}
\label{sec:results}

In this section we first analyze our approach on the MPII-MD \cite{rohrbach15cvpr} dataset and explore different design choices. Then, we compare our best system to prior work.

\subsection{Analysis of our approach}

\paragraph{Experimental setup.}
We build on the labels discovered by our semantic parser \cite{rohrbach15cvpr} and additionally match these labels to sentences which the parser failed to process. To be able to learn classifiers we select the labels that appear at least 30 times, resulting in 1,263 labels. The parser additionally tells us whether the label is a verb. We use the visual features (DT, LSDA, PLACES) provided with the MPII-MD dataset \cite{rohrbach15cvpr}. The LSTM output/hidden unit as well as memory cell have each 500 dimensions.  We train the SVM classifiers on the Training set (56,861 clips). We evaluate our method on the validation set (4,930 clips) using the METEOR \cite{lavie2014meteor} score, which, according to \cite{elliott2013image,vedantam2014cider}, supersedes other popular measures, such as BLEU \cite{papineni02acl}, ROUGE \cite{lin2004rouge}, in terms of agreement with human judgments. 
The authors of CIDEr~\cite{vedantam2014cider} showed that METEOR also outperforms CIDEr when the number of references is small and in the case of MPII-MD we have typically only a single reference.

\newcommand{\midruleValLong}{\cmidrule(rr){1-1} \cmidrule(rr){2-4}}

\begin{table}[t]
\begin{center}
\small
\begin{tabular}{l@{\ \ \ }r@{\ \ \ }r@{\ \ \ }r}
\toprule
         &        & \multicolumn{2}{c}{Classifiers} \\
Approach & Labels & Retrieved   & Trained \\
\midruleValLong
\multicolumn{4}{l}{\textbf{Baseline: all labels treated the same way}} \\
(1) DT                 & 1263 & - & 6.73 \\
(2) LSDA               & 1263 & - & 7.07 \\
(3) PLACES             & 1263 & - & 7.10 \\
(4) DT+LSDA+PLACES     & 1263 & - & 7.24 \\
\multicolumn{4}{l}{\textbf{Visual labels}} \\
(5) Verbs(DT), Others(LSDA)                        & 1328 & 7.08 & 7.27 \\
(6) Verbs(DT), Places(PLACES), Others(LSDA)        & 1328 & 7.09 & 7.39 \\
(7) Verbs(DT), Places(PLACES), Objects(LSDA)       & 913 & 7.10 & 7.48 \\
(8) \ \ + restriction to labels~with~$ROC~\ge~0.7$  & 263 & 7.41 & \textbf{7.54} \\
\multicolumn{4}{l}{\textbf{Baseline: all labels treated the same way, labels from (8)}} \\
(9) DT+LSDA+PLACES                                 & 263 & 7.16 & 7.20 \\
\bottomrule
\end{tabular}
\end{center}
\caption{Comparison of different choices of labels and visual classifiers. All results reported on the validation set of MPII-MD.}
\vspace{-0.5cm}
\label{tbl:valset_labels_viscls}
\end{table}

\subsubsection{Robust visual classifiers.}
In a first set of experiments we analyze our proposal to consider groups of labels to learn different classifiers and also to use different visual representations for these groups (see \secref{sec:visual_labels}). \Tableref{tbl:valset_labels_viscls} we evaluate our generated sentences using different input features to the LSTM. In our baseline, in the top part of \Tableref{tbl:valset_labels_viscls}, we treat all labels equally, \ie we use the same visual descriptors for all labels. The PLACES feature is best with 7.1 METEOR. Combination by stacking all features (DT + LSDA + PLACES) improves further to 7.24 METEOR.

The second part of the table demonstrates the effect of introducing different semantic label groups. We first split the labels into ``Verbs'' and all remaining. Given that some labels appear in both roles, the total number of labels increases to 1328. We analyze two settings of training the classifiers. In the case of ``Retrieved'' we retrieve the classifier scores from the general classifiers trained in the previous step. ``Trained'' corresponds to training the SVMs specifically for each label type (\eg for ``verbs''). Next, we further divide the non-verbal labels into ``Places" and ``Others", and finally into ``Places" and ``Objects". We discard the unused labels and end up with 913 labels. Out of these labels, we select the labels where the classifier obtains a ROC higher or equal to 0.7 (threshold selected on the validation set). After this we obtain 263 labels and the best performance in the ``Trained'' setting. To support our intuition about the importance of the label discrimination (\ie using different features for different semantic groups of labels), we propose another baseline (last line in the table). Here we use the same set of 263 labels but provide the same feature for all of them, namely the best performing combination DT + LSDA + PLACES. As we see, this results in an inferior performance.

We make several observations from \Tableref{tbl:valset_labels_viscls} which lead to robust visual classifiers from the weak sentence annotations. a) It is beneficial to select features based on the label semantics. b) Training one-vs-all SVMs for specific label groups consistently improves the performance as it avoids ``wrong'' negatives. c) Focusing on more ``visual'' labels helps: we reduce the LSTM input dimensionality to 263 while improving the performance.

\subsubsection{LSTM architectures.}
Now, as described in \secref{sec:lstm}, we look at different LSTM architectures and training configurations. In the following we use the best performing ``Visual Labels'' approach,  \Tableref{tbl:valset_labels_viscls} (8).

\newcommand{\midruleValShort}{\cmidrule(lr){1-1} \cmidrule(lr){2-2}}
\begin{table}[t]
\scriptsize
\begin{center}
\begin{subtable}[b]{0.3\textwidth}
\begin{tabular}{lr}
\toprule
Architecture & \scriptsize{$METEOR$} \\
\midruleValShort
{1 layer}             & \textbf{7.54} \\
{2 layers unfact.} & \textbf{7.54} \\
{2 layers fact.}   & {7.41} \\
\bottomrule
\end{tabular}
\caption{LSTM architectures.}
\label{tbl:valset_layers}
\end{subtable}\ \ \ 
\begin{subtable}[b]{0.27\textwidth}
\begin{tabular}{lr}
\toprule
Dropout & \scriptsize{$METEOR$} \\
\midruleValShort
{no dropout}               & 7.19 \\
{lang-drop} & 7.13 \\
{vis-drop}   & 7.34 \\
{concat-drop}           & 7.29 \\
{lstm-drop}   & \textbf{7.54} \\
\bottomrule
\end{tabular}
\caption{Dropout strategies.}
\label{tbl:valset_dropout}
\end{subtable}\ \ \ 
\begin{subtable}[b]{0.3\textwidth}
\begin{tabular}{lr}
\toprule
Dropout ratio & \scriptsize{$METEOR$} \\
\midruleValShort
{r=0.1}         & 7.22 \\
{r=0.25}        & 7.42 \\
{r=0.5}         & \textbf{7.54} \\
{r=0.75}        & 7.46 \\
\bottomrule
\end{tabular}
\caption{Dropout ratios.}
\label{tbl:valset_dropout_value}
\end{subtable}
\caption{(a) Different LSTM architectures, used lstm-dropout 0.5. (b) Comparison of different dropout strategies in 1-layer LSTM with dropout value=0.5. (c) Comparison of different dropout ratios in 1-layer LSTM with lstm-dropout. Labels and classifiers from Table \ref{tbl:valset_labels_viscls} (8). On  validation set of MPII-MD.}
\vspace{-0.7cm}
\end{center}
\end{table}

We start with examining the architecture, where we explore different configurations of LSTM and dropout layers. \Tableref{tbl:valset_layers} shows the performance of three different networks: ``1 layer'', ``2 layers unfactored'' and ``2 layers factored'' introduced in \secref{sec:lstm}. As we see, the ``1 layer'' and ``2 layers unfactored'' perform equally well, while ``2 layers factored'' is inferior to them. In following experiments we use the simplest ``1 layer'' network. We then compare different dropout placements as illustrated in (\Figref{tbl:valset_dropout}). We obtain the best result when applying dropout after the LSTM layer (``lstm-drop''), while having no dropout or applying it only to language leads to stronger over-fitting to the visual features. Putting dropout after the LSTM (and prior to a final prediction layer) makes the entire system more robust. As for the best dropout ratio, we find that 0.5 works best with lstm-dropout \Tableref{tbl:valset_dropout_value}.

\begin{table}[t]
\scriptsize
\begin{center}
\begin{subtable}[b]{0.49\textwidth}
\begin{tabular}{lr}
\toprule
Approach & $METEOR$ \\
\midruleValShort
{lr=0.005, step=2000}  & 7.30 \\
{lr=0.01, step=2000}   & 7.54 \\
{lr=0.02, step=2000}   & 7.51 \\
\midruleValShort
{lr=0.005, step=4000}  & 7.49 \\
{lr=0.01, step=4000}   & \textbf{7.59} \\
{lr=0.02, step=4000}   & 7.28 \\
\midruleValShort
{lr=0.005, step=6000}  & 7.40 \\
{lr=0.01, step=6000}   & 7.40 \\
{lr=0.02, step=6000}   & 7.32 \\
\bottomrule
\end{tabular}
\caption{Base learning rates}
\label{tbl:valset_learning_rate}
\end{subtable}
\begin{subtable}[b]{0.49\textwidth}
\begin{tabular}{lr}
\toprule
Approach & $METEOR$ \\
\midruleValShort
{step=2000, iter=25,000}   & 7.54 \\
{step=4000, iter=25,000}   & \textbf{7.59} \\
{step=6000, iter=25,000}   & 7.40 \\
{step=8000, iter=25,000}   & 7.32 \\
\midruleValShort
{poly, pow=0.5, maxiter=25,000}   & 7.36 \\
{poly, pow=0.5, maxiter=10,000}   & 7.45 \\
{poly, pow=0.7, maxiter=25,000}   & 7.43 \\
{poly, pow=0.7, maxiter=10,000}   & 7.43 \\
\bottomrule
\end{tabular}
\caption{Learning strategies with lr=0.01.}
\label{tbl:valset_learning}
\end{subtable}
\caption{(a) Comparison of different base learning rates, network trained for 25,000 iterations. (b) Comparison of different learning strategies with base lr=0.01. All results reported on the validation Set of MPII-MD.}
\end{center}
\end{table}

We compare different learning rates and learning strategies in \Tablesref{tbl:valset_learning_rate} and \ref{tbl:valset_learning}. We find that the best learning rate in the step-based learning is 0.01, while step 4000 slightly improves over step 2000 (which we used in \Tableref{tbl:valset_labels_viscls}). We explore an alternative learning strategy, namely decreasing learning rate according to a polynomial decay. We experiment with different exponents (0.5 and 0.7) and numbers of iterations (25K and 10K), using the base-learning rate 0.01. Our results show that the step-based learning is superior to the polynomial learning.

\begin{table}[t]
\begin{center}
\scriptsize
\begin{tabular}{lr}
\toprule
Approach & $METEOR$ \\
\midruleValShort
{1 net: lr 0.01, step 2000, iter=25,000} & {7.54} \\
{ensemble of 3 nets}                 & {7.52} \\
\midruleValShort
{1 net: lr 0.01, step 4000, iter=25,000} & {7.59} \\
{ensemble of 3 nets}                 & {7.68} \\
\midruleValShort
{1 net: lr 0.01, step 4000, iter=15,000} & {7.55} \\
{ensemble of 3 nets}                 & \textbf{7.72} \\
\bottomrule
\end{tabular}
\end{center}
\caption{Ensembles of networks with different random initializations. All results reported on the validation set of MPII-MD.}
\label{tbl:valset_rand}
\end{table}

In most of experiments we trained our networks for 25,000 iterations. After looking at the METEOR performance for intermediate iterations we found that for the step size 4000 at iteration 15,000 we achieve best performance overall. Additionally we train multiple LSTMs with different random orderings of the training data. In our experiments we combine three in an ensemble, averaging the resulting word predictions. In most cases the ensemble improves over the single networks in terms of METEOR score (see \Tableref{tbl:valset_rand}).

To summarize, the most important aspects that decrease over-fitting and lead to a better sentence generation are: (a) a correct learning rate and step size, (b) dropout after the LSTM layer, (c) choosing the training iteration based on METEOR score as opposed to only looking at the LSTM accuracy/loss which can be misleading, and (d) building ensembles of multiple networks with different random initializations. In the following section we evaluate our best ensemble (last line of \Tableref{tbl:valset_rand}) on the test set of MPII-MD.

\subsection{Comparison to related work}

\paragraph{Experimental setup.}
We compare the best method of \cite{rohrbach15cvpr}, the recently proposed method S2VT \cite{venugopalan15arxiv} and our proposed ``Visual Labels''-LSTM on the test set of the MPII-MD dataset (6,578 clips). 
We report all popular automatic evaluation measures, CIDEr \cite{vedantam2014cider}, BLEU \cite{papineni02acl}, ROUGE \cite{lin2004rouge} and METEOR \cite{lavie2014meteor}, computed using the evaluation code of \cite{capeval2015}. We also perform a human evaluation, by randomly selecting {1300} video snippets and asking AMT turkers to rank three systems (the best SMT of \cite{rohrbach15cvpr}, S2VT \cite{venugopalan15arxiv} and ours) with respect to Correctness, Grammar and Relevance, similar to \cite{rohrbach15cvpr}.

\newcommand{\midruleTestLong}{\cmidrule(rr){1-1} \cmidrule(rr){2-5} \cmidrule(rr){6-8}}
\newcommand{\rot}[1]{\multicolumn{1}{c}{\begin{turn}{90}\hspace{-2pt}#1\end{turn}}}
\begin{table}[t]
\begin{center}
\footnotesize
\begin{tabular}{l@{\ \ \ }cccc@{\ \ \ }ccc}%
\toprule
         & \multicolumn{4}{c}{Automatic Score}             & \multicolumn{3}{c}{Human evaluation: rank} \\
Approach & \scriptsize{$CIDEr$} & \scriptsize{$BLEU@4$} & \scriptsize{$ROUGE_L$} & \scriptsize{$METEOR$} & {Correct.} & {Grammar} & {Relev.} \\
\midruleTestLong
Best SMT of \cite{rohrbach15cvpr} & 8.14 & 0.47 & 13.21 & 5.59 & 2.11 & 2.39 & 2.08 \\
S2VT \cite{venugopalan15arxiv}    & 9.00 & 0.49 & 15.32 & 6.27 & 2.02 & \textbf{1.67} & 2.06 \\
Our & \textbf{9.98} & \textbf{0.80} & \textbf{16.02} & \textbf{7.03} & \textbf{1.87} & 1.94 & \textbf{1.86} \\
\midruleTestLong
NN Upperbound                         & 169.64 & 9.42 & 44.04 & 19.43 & -& -& -\\
\bottomrule 
\end{tabular}
\end{center}
\caption{Comparison of prior work and our proposed method using all popular evaluation measures. Human scores in form of ranking from 1 to 3, where lower is better. All results reported on the test Set of MPII-MD.}
\vspace{-0.5cm}
\label{tbl:testset}
\end{table}

\paragraph{Results.}
\Tableref{tbl:testset} summarizes the results on the test set of MPII-MD. While we rely on identical features and similar labels as \cite{rohrbach15cvpr}, we significantly improve the performance in all automatic measures, specifically by 1.44 METEOR points. Moreover, we improve over the recent approach of \cite{venugopalan15arxiv}, which also uses LSTM to generate video descriptions. Exploring different strategies to label selection and classifier training, as well as various LSTM configurations allows to obtain best result to date on the MPII-MD dataset.
Human evaluation mainly agrees with the automatic measures. We outperform both prior works in terms of Correctness and Relevance, however we lose to S2VT in terms of Grammar. This is due to the fact that S2VT produces overall shorter (7.4 versus 8.7 words per sentence) and simpler sentences, while our system generates longer sentences and therefore has higher chances to make mistakes. 

We also propose a retrieval upperbound (last line in \Tableref{tbl:testset}). For every test sentence we retrieve the closest training sentence according to the METEOR. The rather low METEOR score of 19.43 reflects the difficulty of the dataset.

\begin{figure}[t]
\scriptsize
\center
\begin{tabular}{l@{\ \ \ }l@{\ \ \ }l}
\toprule
& Approach &  Sentence\\ %
\midrule
\multirow{6}{*}{\includegraphics[width=3cm]{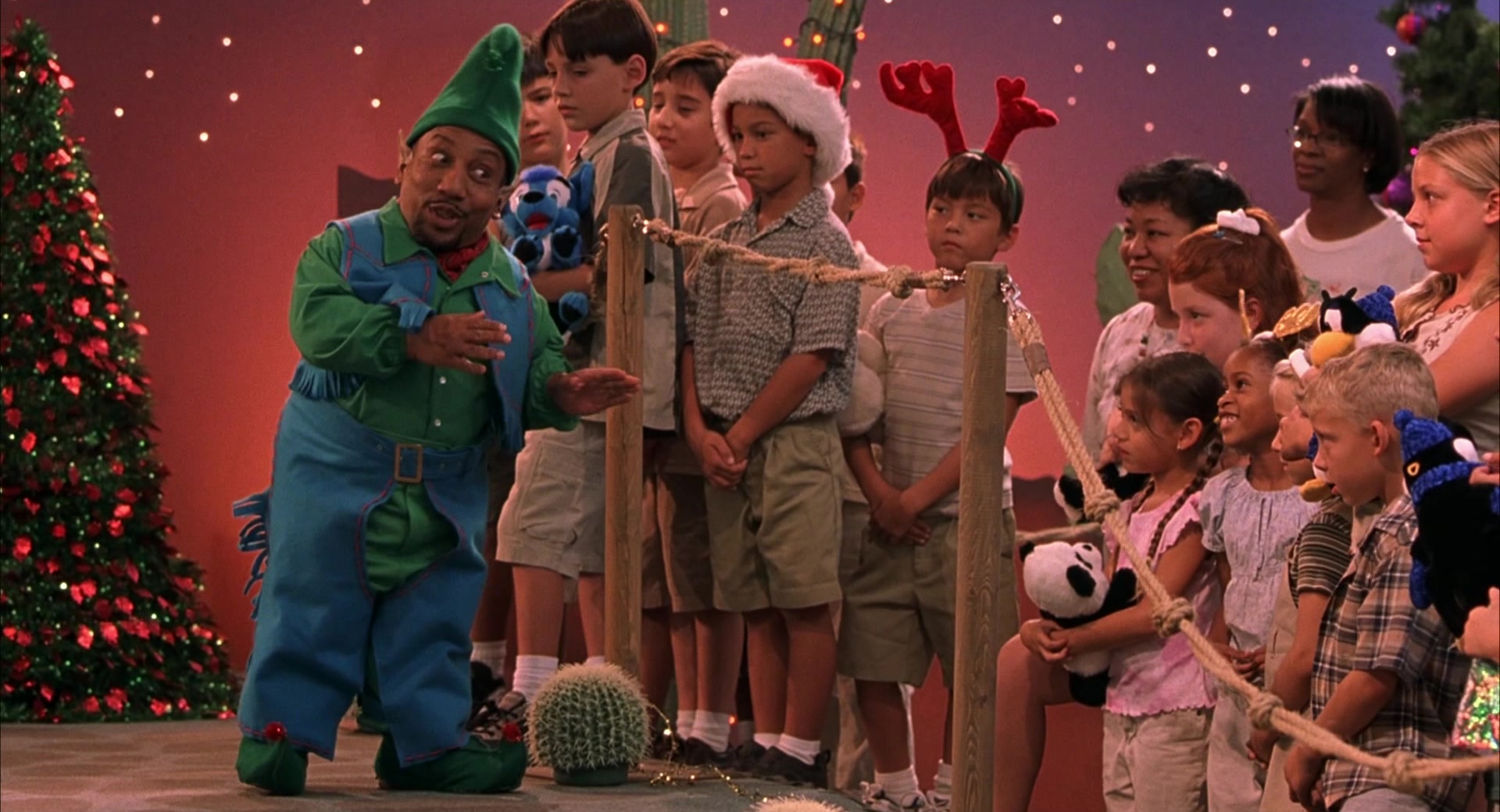}} & SMT \cite{rohrbach15cvpr} & Someone is a man, someone is a man. \\ 
 &S2VT \cite{venugopalan15arxiv} & Someone looks at him, someone turns to someone.\\ 
 &Our & Someone is standing in the crowd, \\
 &&a little man with a little smile. \\
 & Reference & Someone, back in elf guise, is trying to calm the kids. \\
 \\
\multirow{6}{*}{\includegraphics[width=3cm]{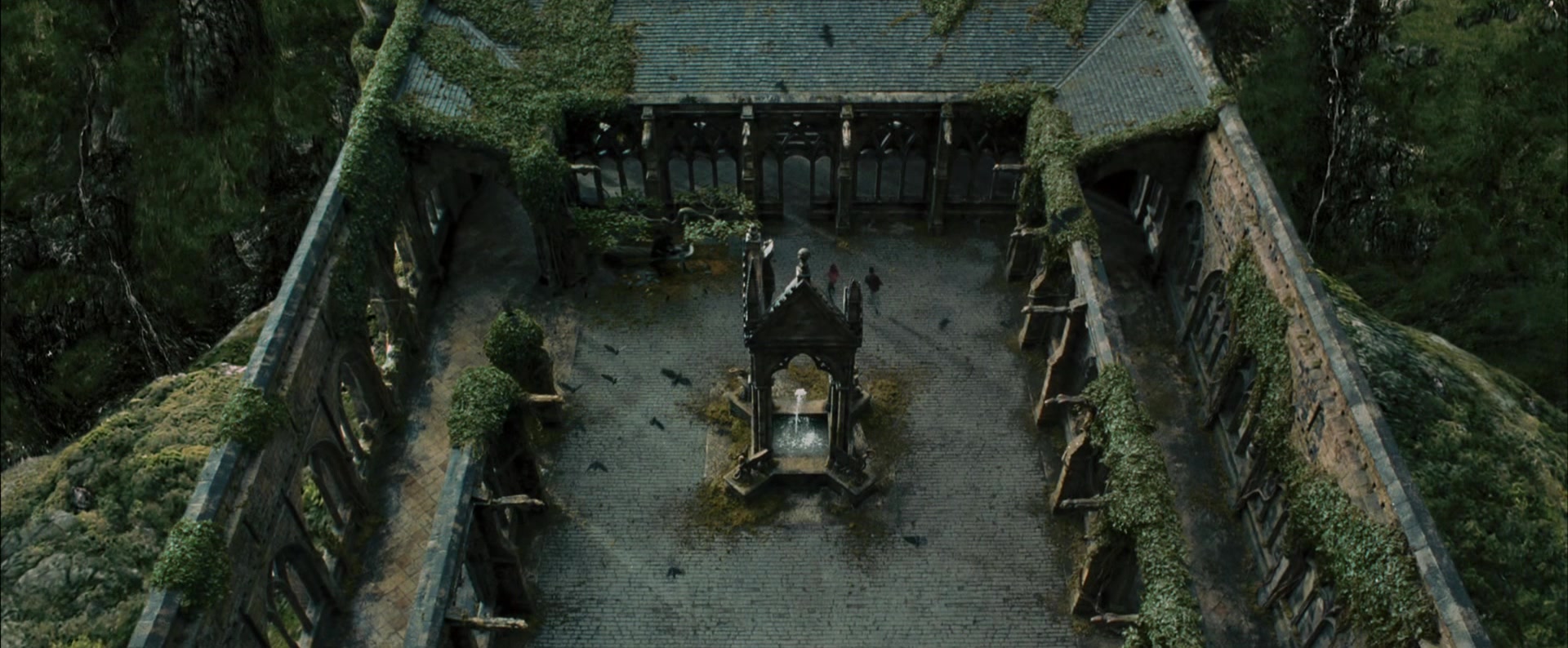}} &  \\
& SMT \cite{rohrbach15cvpr} &The car is a water of the water. \\
 &S2VT \cite{venugopalan15arxiv} &  On the door, opens the door opens. \\
  &Our &  The fellowship are in the courtyard. \\
  &Reference & They cross the quadrangle below and run along the cloister. \\
  \\
\multirow{6}{*}{\includegraphics[width=3cm]{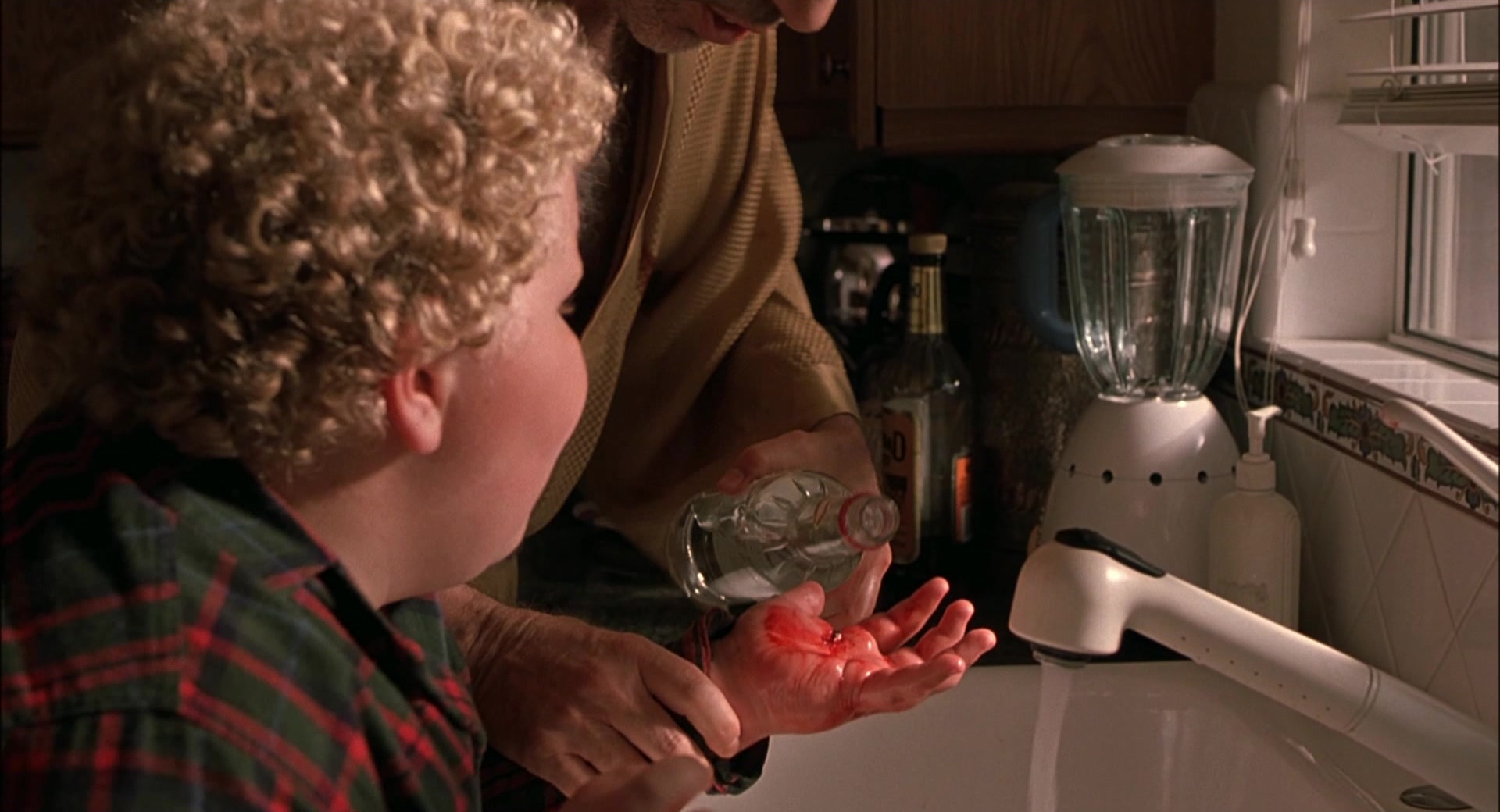}}& SMT \cite{rohrbach15cvpr} & Someone is down the door, \\
&&someone is a back of the door, and someone is a door. \\
 &S2VT \cite{venugopalan15arxiv} &   Someone shakes his head and looks at someone.\\ 
 &Our & Someone takes a drink and pours it into the water. \\
 &Reference& Someone grabs a vodka bottle standing open on the counter \\
 && and liberally pours some on the hand. \\
\bottomrule
\end{tabular}
\caption{Qualitative comparison of prior work and our proposed method. Examples from the test set of MPII-MD. Our approach identifies activities, objects, and places better than related work.}
\label{fig:qual}
\end{figure}

A closer look at the sentences produced by all three methods gives us additional insights. An interesting characteristic is the output vocabulary size, which is \emph{94} for \cite{rohrbach15cvpr}, \emph{86} for \cite{venugopalan15arxiv} and \emph{605} for our method, while the test set contains \emph{6422} unique words. This clearly shows a higher diversity of our output. Among the words generated by our system and absent in the outputs of others are such verbs as \emph{grab, drive, sip, climb, follow}, objects as \emph{suit, chair, cigarette, mirror, bottle} and places as \emph{kitchen, corridor, restaurant}. We showcase some qualitative results in \Figref{fig:qual}. Here, e.g. the verb \emph{pour}, object \emph{drink} and place \emph{courtyard} only appear in our output. We attribute this, on one hand, to our diverse and robust visual classifiers. On the other hand, the architecture and parameter choices of our LSTM allow us to learn better correspondance between words and visual classifiers' scores.

\section{Analysis}
\label{sec:analysis}

\begin{figure}[t!]
\begin{center}
\begin{subfigure}[t]{0.4\textwidth}
\includegraphics[width=5cm]{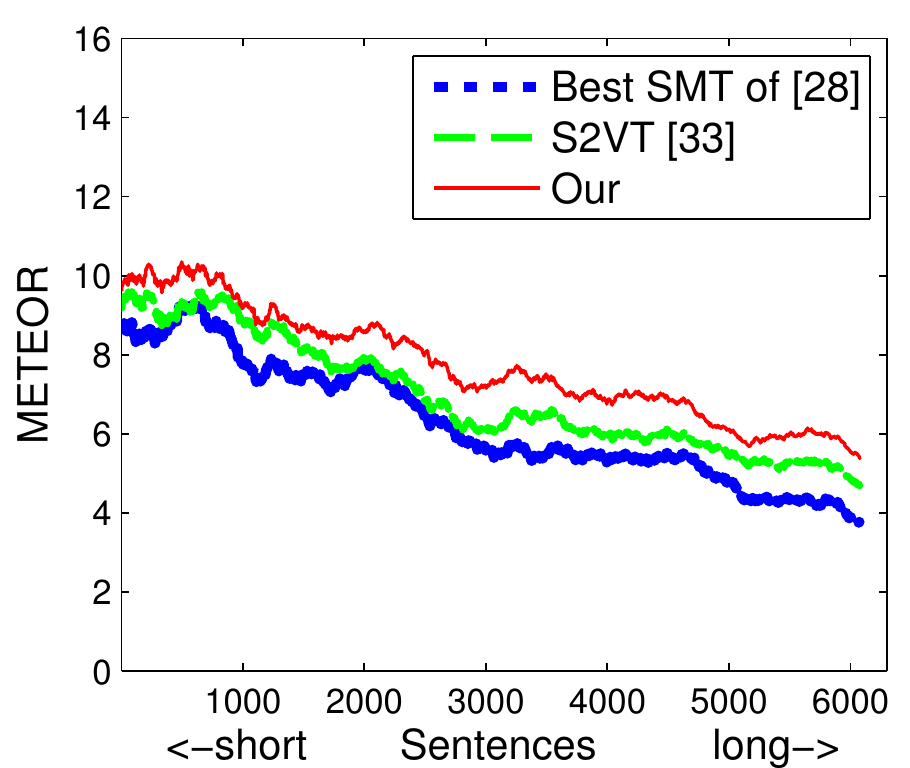}\\
\vspace{-0.3cm}
\caption{Sentence length}
\label{fig:sent_length}
\end{subfigure}
\begin{subfigure}[t]{0.4\textwidth}
\includegraphics[width=5cm]{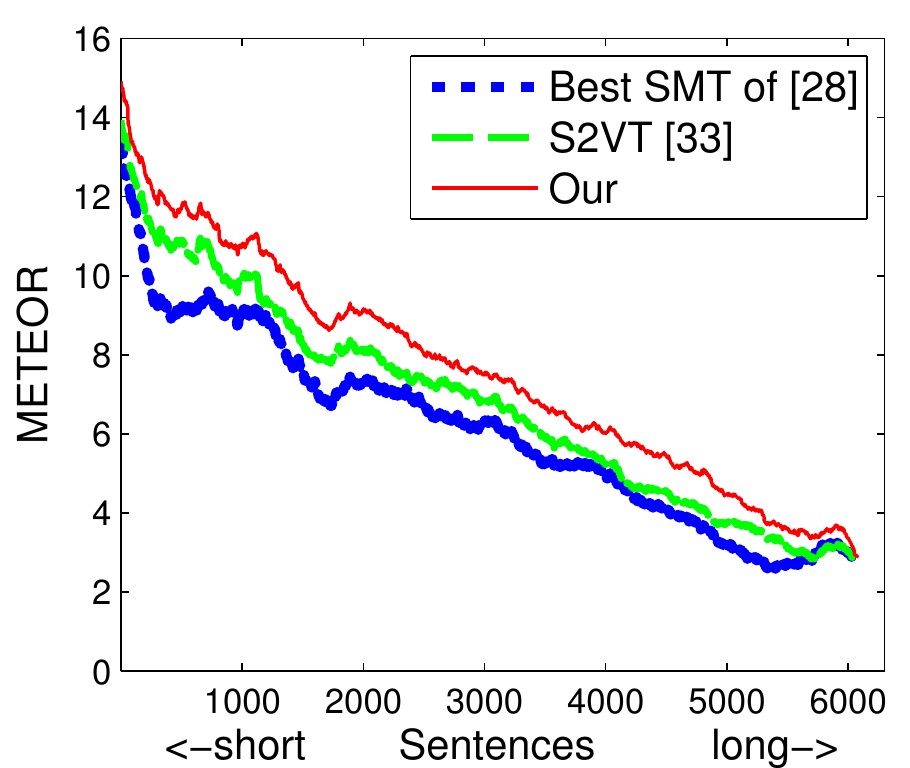}\\
\vspace{-0.3cm}
\caption{Word frequency}
\label{fig:word_freq}
\end{subfigure}
\end{center}
\vspace{-0.5cm}
\caption{METEOR score per sentence. (a) Test set sorted by sentence length (increasing). (b) Test set sorted by word frequency (decreasing). Shown values are smoothed with a mean filter of size 500. }
\label{fig:difficulty1}
\end{figure}

Despite the recent advances in the video description domain, including our proposed approach, the video description performance on the movie description datasets (MPII-MD \cite{rohrbach15cvpr} and M-VAD \cite{torabi15arxiv}) remains relatively low. In this section we want to take a closer look at three methods, best SMT of \cite{rohrbach15cvpr}, S2VT \cite{venugopalan15arxiv} and ours, in order to understand where these methods succeed and where they fail. In the following we evaluate all three methods on the MPII-MD test set.
\subsection{Difficulty versus performance}
As the first study we suggest to sort the reference sentences (from the test set) by difficulty, where difficulty is defined in multiple ways.

\textbf{Sentence length and Word frequency.}
Two of the simplest sentence difficulty measures are its length and average frequency of words. When sorting the data by difficulty (increasing sentence length or decreasing average word frequency), we find that all three methods have the same tendency to obtain lower METEOR score as the difficulty increases (\Figsref{fig:sent_length} and \ref{fig:word_freq}). For the word frequency the correlation is stronger. Our method consistently outperforms the other two, most notable as the difficulty increases.

\begin{figure}[t!]
\begin{center}
\begin{subfigure}[t]{0.4\textwidth}
\includegraphics[width=5cm]{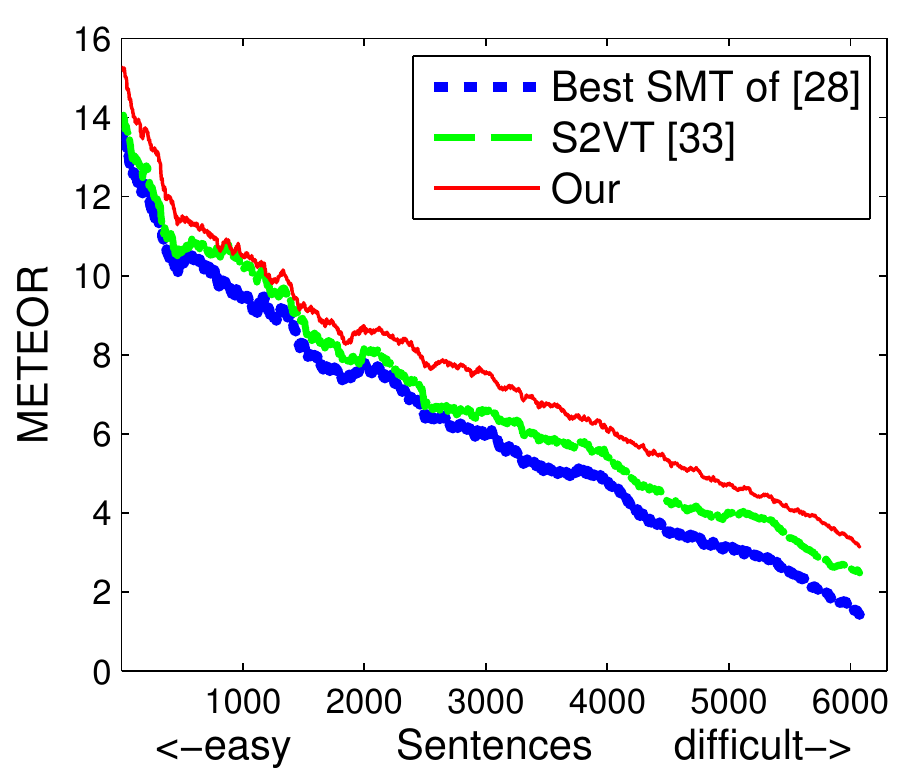}\\
\vspace{-0.3cm}
\caption{Textual dificulty}
\label{fig:tnn}
\end{subfigure}
\begin{subfigure}[t]{0.4\textwidth}
\includegraphics[width=5cm]{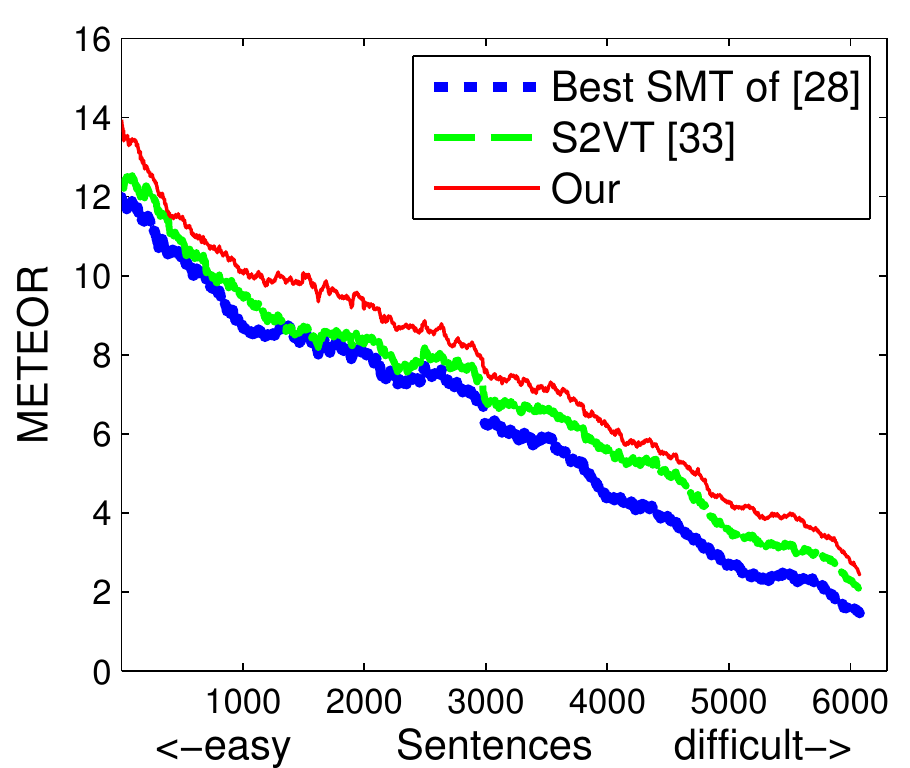}\\
\vspace{-0.3cm}
\caption{Visual difficulty}
\label{fig:vnn}
\end{subfigure}
\end{center}
\vspace{-0.5cm}
\caption{METEOR score per sentence. (a) Test set sorted by Textual NN score (decreasing). (b) Test set sorted by Visual kNN score, $k=10$ (decreasing). Shown values are smoothed with a mean filter of size 500. }
\label{fig:difficulty2}
\end{figure}

\textbf{Textual and Visual Nearest Neighbors.}
Next, for each reference test sentence we search for the closest training sentence (in terms of the METEOR score). We use the obtained best scores to sort the reference sentences by textual difficulty, \ie the ``easy'' sentences are more likely to be retrieved. If we consider all training sentences, we obtain a Textual Nearest Neighbor. We sort the test sentences according to these scores (decreasing) and plot the performance of three methods in \Figref{fig:tnn}. All methods ``agree'' and ours is best throughout the difficulty range, in particular in the more challenging part of the plot. We can also use visual features to find the $k$ Nearest Neighbors in the Training set, select the best one (in terms of the METEOR score) and use this score to sort the reference sentences. We call this a Visual k Nearest Neighbor. The intuition behind it is to consider a video clip as visually ``easy'' if the most similar training clips also have similar descriptions (the ``difficult'' clip might have no close visual neighbours).
We rely on our best visual representation (8) from \Tableref{tbl:valset_labels_viscls} and $cos$ similarity measure to define the Visual kNN and sort the reference sentences according to it with $k=10$ (\Figref{fig:vnn}). We see a clear correlation between the visual difficulty and the performance of all methods (\Figref{fig:vnn}).

\textbf{Summary.} a) All methods perform better on shorter, common sentences and our method notably wins on longer sentences. b) Our method also wins on sentences that are more difficult to retrieve. 
c) Visual difficulty, defined by $cos$ similarity and representation (8) from \Tableref{tbl:valset_labels_viscls}, strongly correlates with the performance of all methods. (d) When comparing all four plots (\Figsref{fig:sent_length} and \ref{fig:word_freq}, \Figsref{fig:tnn} and \ref{fig:vnn}), we find that the strongest correlation between the methods' performance and the difficulty is observed for the Textual difficulty, while the least correlation we observe for the Sentence length.

\subsection{Semantic analysis}

\begin{figure}[t]
\begin{center}
\includegraphics[width=0.9\linewidth]{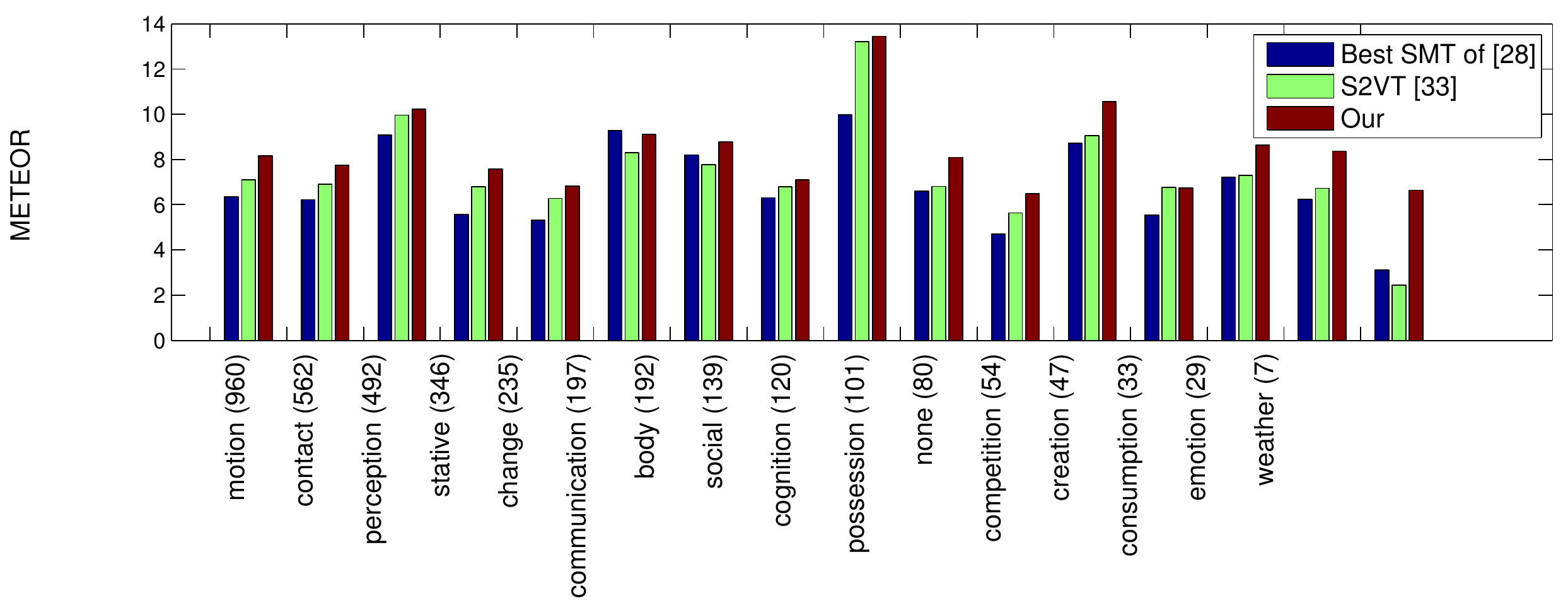}
\caption{Average METEOR score for WordNet verb topics. Selected sentences with single verb, number of sentences in brackets.}
\label{fig:topics}
\end{center}
\end{figure}

\begin{table}[t]
\begin{center}
\scriptsize
\begin{tabular}{lllllll}
Topic & Entropy & Top-1 & Top-2 & Top-3 & Top-4 & Top-5 \\
\toprule
\textbf{motion} & 7.05 & turn & walk & shake & move & go \\
\textbf{contact} & 7.10 & open & sit & stand & hold & pull \\
\textbf{perception} & 4.83 & look & stare & see & watch & gaze \\
\textbf{stative} & 4.84 & be & follow & stop & go & wait \\
\textbf{change} & 6.92 & reveal & start & emerge & fill & make \\
\textbf{communication} & 6.73 & look up & nod & face & speak & talk \\
\textbf{body} & 5.04 & smile & wear & dress & grin & glare \\
\textbf{social} & 6.11 & watch & join & do & close & make \\
\textbf{cognition} & 5.21 & look at & see & read & take & leave \\
\textbf{possession} & 5.29 & give & take & have & stand in & find \\
\textbf{none} & 5.04 & throw & hold & fly & lie & rush \\
\textbf{creation} & 5.69 & hit & make & do & walk through & come up \\
\textbf{competition} & 5.19 & drive & walk over & point & play & face \\
\textbf{consumption} & 4.52 & use & drink & eat & take & sip \\
\textbf{emotion} & 6.19 & draw & startle & feel & touch & enjoy \\
\textbf{weather} & 3.93 & shine & blaze & light up & drench & blow \\
\bottomrule
\end{tabular}
\end{center}
\caption{Entropy and top 5 frequent verbs of each WordNet topic in the MPII-MD.}
\label{tbl:topic_words}
\end{table}

\textbf{WordNet Verb Topics.}
We closer analyze the test sentences with respect to different verbs. For this we rely on WordNet topics (high level entries in the WordNet ontology, \eg ``motion'', ``perception'', ``competition'', ``emotion''), defined for most synsets in WordNet \cite{fellbaum:wordnet}. We obtain the sense information from the semantic parser of \cite{rohrbach15cvpr}, thus senses might be noisy. We showcase the 5 most frequent verbs for each topic in \Tableref{tbl:topic_words}. We select sentences with a single verb, group them according to the verb topic and compute an average METEOR score for each topic, see \Figref{fig:topics}. We find that our method is best for all topics except ``communication", where \cite{rohrbach15cvpr} wins. The most frequent verbs in this topic are ``look up'' and ``nod'', which are also frequent in the dataset and in the sentences produced by \cite{rohrbach15cvpr}. The best performing topic, ``cognition'', is highly biased to ``look at'' verb. The most frequent topics, ``motion'' and ``contact'', which are also visual (\eg ``turn'', ``walk'', ``open'', ``sit''), are nevertheless quite challenging, which we attribute to their high diversity (see their entropy w.r.t. different verbs and their frequencies in \Tableref{tbl:topic_words}). At the same time ``perception'' is far less diverse and mainly focuses on verbs like ``look'' or ``stare'', which are quite frequent in the dataset, resulting in better performance. Topics with more abstract verbs (\eg ``be'', ``have'', ``start'') tend to get lower scores.

\textbf{Top 100 best and worst sentences.}
We look at 100 Test sentences, where our method obtains highest and lowest METEOR scores. Out of 100 best sentences 44 contain the verb ``look'' (including verb phrases such as ``look at''). The other frequent verbs are ``walk'', ``turn", ``smile'', ``nod'', ``shake'', ``stare'', ``sit'', \ie mainly visual verbs. Overall the sentences are simple and common. Among the 100 lowest scoring sentences we observe more diversity: 12 sentences contain no verb, 10 mention unusual words (specific to the movie), 24 contain no subject, 29 have a non-human subject. Altogether this leads to a lower performance, in particular, as most training sentences contain ``Someone'' as subject and generated sentences are biased towards it.

\textbf{Summary.} a) The test sentences that mention the verb ``look'' (and similar) get higher METEOR scores due to their high frequency in the dataset. b) The sentences with more ``visual'' verbs tend to get higher scores. c) The sentences without verbs (\eg describing a scene), without subjects or with non-human subjects get lower scores, which can be explained by a dataset bias towards ``Someone'' as subject.

\section{Conclusion}
\label{sec:conclusion}
We propose an approach to automatic movie description which trains visual classifiers and uses the classifier scores as input to LSTM. To handle the weak sentence annotations we rely on three main ingredients. First, we distinguish three semantic groups of labels (verbs, objects and places), second we train them discriminatively, removing potentially noisy negatives, and third, we select only a small number of the most reliable classifiers.
For sentence generation we show the benefits of exploring different LSTM architectures and learning configurations. As the result we obtain the highest performance on the MPII-MD dataset as shown by all automatic evaluation measures and extensive human evaluation.

We analyze the challenges in the movie description task using our and two prior works. We find that the factors which contribute to higher performance include: presence of frequent words, sentence length and simplicity as well as presence of ``visual'' verbs (\eg ``nod'', ``walk'', ``sit'', ``smile''). Textual and visual difficulties of sentences/clips strongly correlate with the performance of all methods. We observe a high bias in the data towards humans as subjects and verbs similar to ``look''. Future work has to focus on dealing with less frequent words and handle less visual descriptions. This potentially requires to consider external text corpora, modalities other than video, such as audio and dialog, and to look across multiple sentences. This would allow exploiting long- and short-range context and thus understanding and describing the story of the movie.

\paragraph{Acknowledgements.}
Marcus Rohrbach was supported by a fellowship within the FITweltweit-Program of the German Academic Exchange Service (DAAD). The authors thank Niket Tandon for help with the WordNet Topics analysis.

\bibliographystyle{splncs03}
\bibliography{biblioLong,rohrbach,rohrbach15cvpr}

\end{document}